\documentclass[letterpaper, 10 pt, conference]{ieeeconf}
\IEEEoverridecommandlockouts
\title{\LARGE \bf
Robots with Attitude: Singularity-Free Quaternion-Based Model-Predictive Control for Agile Legged Robots
}

\author{Zixin Zhang$^{1}$, John Z. Zhang$^{2}$, Shuo Yang$^{3}$, and Zachary Manchester$^{2}$
\thanks{$^{1}$Zixin Zhang is with the Center for Robotics and Biosystems, Northwestern University, Evanston, IL 60208, USA {\tt\small\ zixinzhang2027@u. northwestern.edu}}
\thanks{$^{2}$John. Z. Zhang and Zachary Manchester are with the Robotics Institute, Carnegie Mellon University, Pittsburgh, PA 15213, USA {\tt\small\{johnzhang, zacm\}@cmu.edu}}
\thanks{$^{2}$Shuo Yang is with the Department of Mechanical Engineering, Carnegie Mellon University, Pittsburgh, PA 15213, USA {\tt\small\ shuoyang@andrew.cmu.edu}}}

\usepackage{graphics}
\usepackage{epsfig}
\usepackage{mathptmx}
\usepackage[mathscr]{euscript}
\usepackage{times}
\usepackage{amssymb}
\usepackage{amsmath}
\usepackage{graphicx}
\usepackage{array}
\usepackage{tabularx}
\usepackage{multirow}
\usepackage{amsfonts} 
\usepackage{booktabs}
\usepackage{xcolor}
\usepackage{subcaption}
\usepackage{adjustbox}
\usepackage{pdfpages}
\usepackage[abs]{overpic}
\usepackage[T1]{fontenc}
\usepackage[utf8]{inputenc}
\usepackage{pgfplots}
\usepackage{grffile}

\pgfplotsset{compat=newest}
\usetikzlibrary{plotmarks}
\usetikzlibrary{arrows.meta}
\usepgfplotslibrary{patchplots}

\usepackage{newtxtext,newtxmath}
\usepackage{pdfrender}

\newcommand*{\boldgreek}[1]{%
  \textpdfrender{%
    TextRenderingMode=FillStroke,%
    LineWidth=.35pt,%
  }{#1}%
}

\makeatletter
\let\NAT@parse\undefined
\makeatother
\usepackage{hyperref}

\begin{document}

\maketitle
\thispagestyle{empty}
\pagestyle{empty}


\begin{abstract}

We present a model-predictive control (MPC) framework for legged robots that avoids the singularities associated with common three-parameter attitude representations like Euler angles during large-angle rotations. Our method parameterizes the robot's attitude with singularity-free unit quaternions and makes modifications to the iterative linear-quadratic regulator (iLQR) algorithm to deal with the resulting geometry. The derivation of our algorithm requires only elementary calculus and linear algebra, deliberately avoiding the abstraction and notation of Lie groups. We demonstrate the performance and computational efficiency of quaternion MPC in several experiments on quadruped and humanoid robots.

\end{abstract}


\section{Introduction}
Legged mobility has been an area of great interest in recent years due to its wide-ranging potential applications, including search and rescue, exploration, and transportation. Legged robots possess the unique ability to traverse complex terrains, navigate through cluttered environments, and perform tasks that are inaccessible to wheeled or tracked robots. A key challenge of legged locomotion is to efficiently control a robot through challenging terrain and perform agile maneuvers in which large changes in attitude can occur.

Current control methods for legged robots commonly represent rotations with Euler angles \cite{Di2018, cleach2021}, which are prone to singularities that can cause controllers to fail. Recent works have explored the use of Lie group representations in model-predictive controllers for quadrupeds~\cite{teng2022} and hoppers~\cite{Aaron2023}, however the exposition and implementation of these algorithms requires a deep understanding of Lie groups and Lie algebras that is not yet widespread in the robotics comunnity.

Our goal is to derive a simple MPC algorithm that offers a globally non-singular 3D rotation state for legged robots. To achieve this, we use unit quaternions to represent the 3D orientation of the robot and utilize only standard linear algebra and basic calculus results to optimize directly in this state space \cite{Jackson2021}. We incorporate the quaternion optimization method into a nonlinear MPC framework with single-rigid-body (SRB) dynamics. Compared to Euler-angle-based MPC (Euler MPC), which suffers from singularities, our method — which we refer to as quaternion MPC — handles arbitrarily large changes in the robot's attitude.

\begin{figure}
    \vspace{2mm}
    \centering\includegraphics[width=1.0\linewidth]{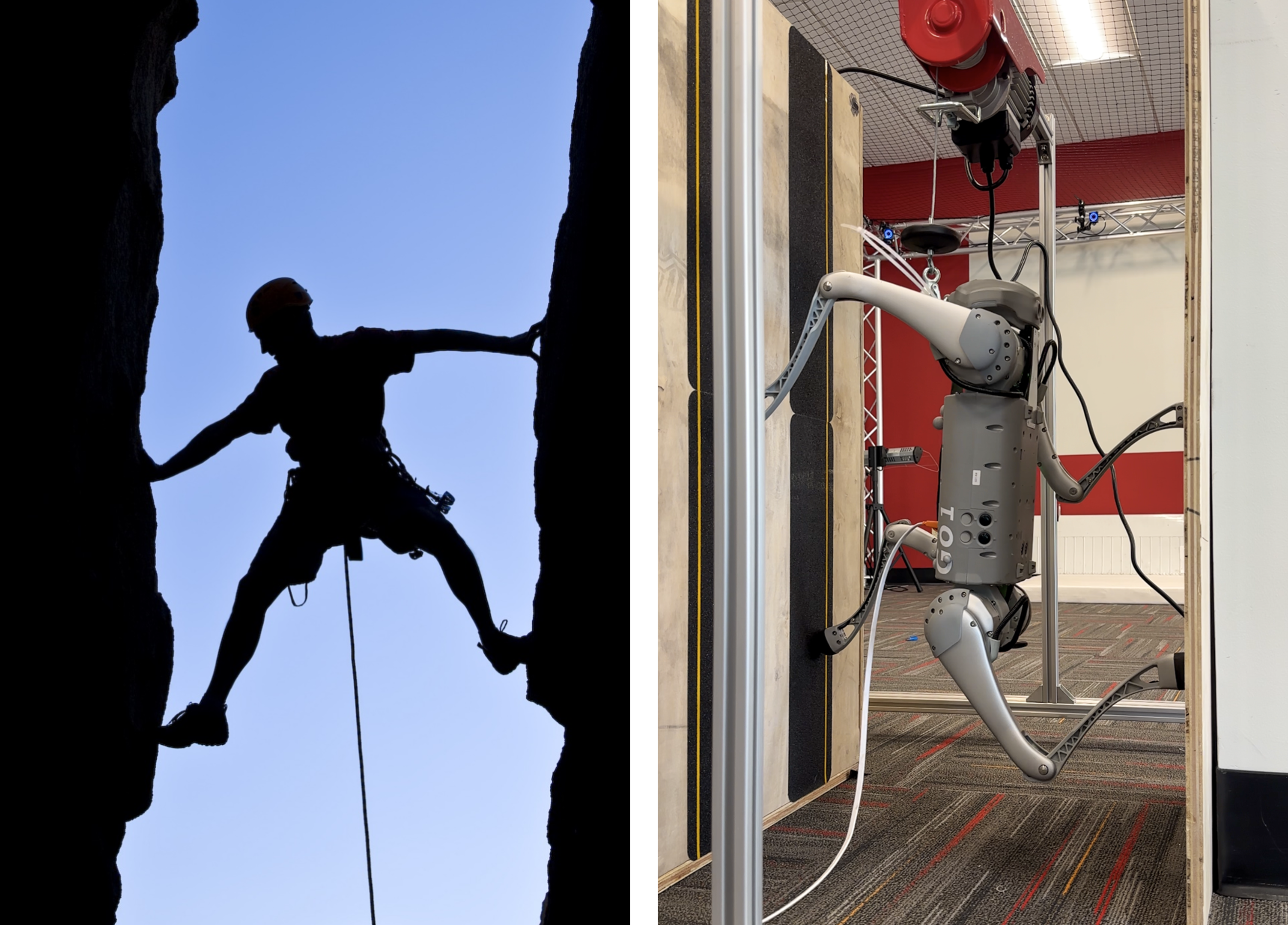}
    \caption{A Unitree Go1 robot standing vertically between two walls using quaternion MPC (right) similar to a human climber \cite{human_climber} (left).
    }
    \label{fig:wall_standing}
    \vspace{-7mm}
\end{figure}

We demonstrate the effectiveness of quaternion MPC through several hardware and simulation experiments, tackling a range of tasks including attitude control, locomotion, and disturbance rejection. Notably, Our method enables a quadruped robot to stand between two walls at a 90-degree pitch angle --- a pose that results in a singularity when using Euler angles (Fig. \ref{fig:wall_standing}). We have also used the method in simulations to control a humanoid robot in such a way that it can achieve almost any pose inside the workspace. In addition, for a quadruped robot with two reaction wheels \cite{Lee2023}, quaternion MPC shows superior performance and stability in airborne attitude control during falling compared to Euler MPC.

Attempting to achieve these tasks solely with Euler angles would necessitate the controller adapting the sequence of Euler angles according to the specific task, resulting in discontinuous or hybrid dynamics models. This poses challenges for motion planning and control, making the process complex and less streamlined. Alternatively, while a rotation matrix could be used, its requirement for nine parameters to represent orientation is excessively redundant.

Our specific contributions include:
\begin{itemize}
    \item A singularity-free MPC formulation for legged robots that directly optimizes over the globally well-defined unit quaternions while relying on only standard linear algebra and basic calculus.
    \item Comparisons between a standard MPC algorithm and quaternion MPC in simulation and on hardware.
    \item An open-source implementation of quaternion MPC for legged robots\footnote{\url{https://github.com/zixinz990/quaternion-mpc}}, including several examples of quadruped and humanoid robot control. It can be readily executed on the Unitree Go1 robot and easily ported to other hardware systems.
    
\end{itemize}

This paper is organized as follows: Sec. \ref{related_works} surveys recent work on MPC for legged robots, with a particular emphasis on rotation representations. Sec. \ref{sec:background} reviews necessary background material on unit quaternions, quaternion differential calculus, and MPC for legged robots. Sec. \ref{sec:quat_mpc} then derives our nonlinear MPC formulation and control architecture. Sec. \ref{sec:results} showcases our experimental results. Finally, Sec. \ref{sec:conclusions} summarizes our conclusions and suggests future research directions.


\section{Related Works}\label{related_works}

In recent years, model-predictive control methods have emerged as a popular approach for controlling agile, dynamic locomotion behaviors \cite{Di2018, cleach2021, Bledt2018, zhang2023}. MPC typically uses heuristics for choosing foothold locations \cite{Raibert1985} and solves a convex quadratic program (QP) \cite{osqp, bishop_relu-qp_2023} for the desired foot-stance forces over some prediction horizon. Despite using a linearized single-rigid-body model to represent the quadruped dynamics, this basic MPC formulation is surprisingly effective, even on uneven terrains \cite{Di2018, Bledt2018}. More recently, nonlinear single-rigid body dynamics have been shown to offer improved performance on challenging terrains \cite{Neunert2018, Grandia2022}.


While locomotion capabilities from prior MPC methods have been impressive, their choice of Euler angles as attitude representation poses some fundamental limitations. Many heuristics have been developed over the years to overcome singularities or ``gimbal lock.'' For example, \cite{Di2018} performs quadruped backflips by first optimizing the trajectory in 2D and computing the corresponding feed-forward torques for the 3D model \cite{Katz2019}. Such heuristics can be effective for specific behaviors on a single robot, but fail to generalize. It is worth mentioning that, although such roll and pitch singularities might seem like edge scenarios for current quadruped robots, they become much more important for dynamic, agile humanoid robot behaviors \cite{chignoli2021mit}.

In this paper, we study a general method for optimizing legged robots with 3D rotations. Due to the complex group structure of 3D rotations, three-parameter representations such as Euler angles, axis-angle vectors, and Rodriguez parameters, fail to capture 3D orientations without singularities. Prior works have used rotation matrices \cite{ding2021representation, ding2022orientation} or other Lie groups \cite{teng2022,Aaron2023} to overcome singularities in the quadruped attitude representation. On the other hand, quaternions \cite{Hamilton1844} use four parameters, which is minimal for smooth, singularity-free 3D rotation representations \cite{Stuelpnagel1964}. While quaternions have been widely deployed on spacecraft~\cite{Wie1985}, quadcopters~\cite{Tayebi2006, Fresk2013, Reyes2013}, and underwater vehicles \cite{Fjellstad1994}, they remain under explored for legged locomotion.

\section{Background}\label{sec:background}

In this section, we provide an overview of quaternions, focusing on the operational rules associated with unit quaternions and the differentiation techniques for functions involving quaternions. Our notation is largely based on \cite{Jackson2021}.

\subsection{Unit Quaternions}


In three-dimensional Euclidean space, a rotation can be represented as a unit quaternion $\textbf{q} \in \mathbb{H}:= \left[q_s\text{ }\textbf{q}_v^\intercal \right]^\intercal $ where $\textbf{q}^T\textbf{q} = 1$ and $q_s$ and $\textbf{q}_v$ are the scalar and vector parts of the quaternion, respectively. Note that the space of unit quaternion is a double cover of the rotation group $SO(3)$, which means quaternions $\textbf{q}$ and $-\textbf{q}$ represent the same rotation.

Quaternion multiplication is non-commutative and can be defined as follows:
\begin{equation}\label{eq:quat_mult}
    \textbf{q}_1\otimes \textbf{q}_2:=L\left(\textbf{q}_1\right)\textbf{q}_2,
\end{equation}
where
\begin{equation}\label{eq:L(q)}
    L\left(\textbf{q}\right):=
    \begin{bmatrix}
        q_s & -\textbf{q}_v^\intercal  \\
        \textbf{q}_v & q_s\textbf{I}_3+\left[\textbf{q}_v\right]
    \end{bmatrix},
\end{equation}
and $\left[\textbf{q}_v\right]$ is the $3\times3$ skew-symmetric matrix corresponding to the cross product with $\textbf{q}_v$ \cite{Jackson2021}.

It is often useful to create a ``pure-vector'' quaternion from a three-dimensional vector $\textbf{v} \in \mathbb{R}^3$:
\begin{equation}\label{eq:vec_quat}
    \hat{\textbf{v}}:=
    \begin{bmatrix}
        0 \\
        \textbf{I}_3
    \end{bmatrix}\textbf{v}\equiv \textbf{H}\textbf{v}.
\end{equation}


\subsection{Quaternion Differential Calculus}
A succinct and efficient technique for computing the Jacobians and Hessians of functions involving quaternions is crucial for solving quaternion optimization problems. We note, first, that differential rotations are vectors in $\mathbb{R}^3$, even when using higher-dimensional global rotation representations. We denote these 3-parameter vectors $\boldgreek{\upphi}$. There are many ways to reconstruct a unit quaternion from $\boldgreek{\upphi}$ \cite{Jackson2021}. We chose the Cayley map, of which the forward form is:
\begin{equation}
    \textbf{q} = \varphi(\boldgreek{\upphi}) =  \frac{1}{\sqrt{1 + | \boldgreek{\upphi} |^2}} \begin{bmatrix}
    1 \\
    \boldgreek{\upphi}
    \end{bmatrix},
\end{equation}
and the inverse is:
\begin{equation}
    \boldgreek{\upphi} = \varphi^{-1}(\textbf{q}) = \frac{\textbf{q}_v}{q_s}.
\end{equation}
The Cayley map is easy to compute, and its Jacobian evaluated at $\boldgreek{\upphi}=\textbf{0}$ is $\partial\varphi/\partial\boldgreek{\upphi}=\textbf{H}$.

For vector-valued functions with quaternion inputs $\textbf{y}=h\left(\textbf{q}\right):\mathbb{H}\rightarrow\mathbb{R}^p$, the Jacobian $\nabla h\left(\textbf{q}\right)\in\mathbb{R}^{p\times3}$ with respect to $\boldgreek{\upphi}$ at $\boldgreek{\upphi}=0$ can be computed as:
\begin{equation}\label{eq:vec_fun_jac}
    \nabla h\left(\textbf{q}\right)=\frac{\partial h}{\partial \textbf{q}}L\left(\textbf{q}\right)\textbf{H}:=\frac{\partial h}{\partial \textbf{q}}G\left(\textbf{q}\right),
\end{equation}
where $G\left(\textbf{q}\right)\in\mathbb{R}^{4\times3}:=L\left(\textbf{q}\right)\textbf{H}$ is the \textit{attitude Jacobian}.

In the case where the aforementioned function is a scalar-valued function such as an objective function, with $p=1$, the Hessian is:
\begin{equation}\label{eq:scalar_fun_hess}
    \nabla^2h\left(\textbf{q}\right)=G\left(\textbf{q}\right)^\intercal \frac{\partial^2h}{\partial \textbf{q}^2}G\left(\textbf{q}\right)-\textbf{I}_3\frac{\partial h}{\partial \textbf{q}}\textbf{q}.
\end{equation}

The calculation of the Jacobian for quaternion-valued functions $\textbf{q}'=f\left(\textbf{q}\right):\mathbb{H}\rightarrow\mathbb{H}$ can be obtained through:
\begin{equation}\label{eq:quat_fun_jac}
    \nabla f\left(\textbf{q}\right)=G\left(\textbf{q}'\right)^\intercal \frac{\partial f}{\partial \textbf{q}}G\left(\textbf{q}\right).
\end{equation}

\subsection{Model-Predictive Control for Legged Robots}


A discrete-time MPC policy iteratively solves the following constrained trajectory optimization problem to compute the controls for a dynamical system given its current state, in a feedback fashion:
\begin{equation}\label{eq:dt_mpc}
    \begin{array}{ll}
        \underset{\textbf{x}_{1:K}, \textbf{u}_{1:K-1}}{\mbox{minimize}} & J = l_K(\textbf{x}_K) + \sum \limits_{k = 1}^{K-1} l_k(\textbf{x}_k, \textbf{u}_k)\\
        \mbox{subject to} & \textbf{x}_{k+1} = f_k(\textbf{x}_k,\textbf{u}_k), \quad k = 1,\dots, K-1,\\
        & g_k(\textbf{x}_k, \textbf{u}_k) \leq 0,\\ 
        & h_k(\textbf{x}_k, \textbf{u}_k) = 0.
    \end{array}
\end{equation}
This optimization problem minimizes the cost $J$ over a planning horizon $K$, subject to the discrete-time dynamics $f_k$ and other general equality constraints $h_k$ and inequality constraints $g_k$. Then, the first control input in the optimal input trajectory is applied to the system and the process is repeated.

A common algorithm for efficiently solving problem \eqref{eq:dt_mpc} is Differential Dynamic Programming \cite{Mayne1966ASG}, also known as the iterative Linear-Quadratic Regulator (iLQR) \cite{Li2004IterativeLQ}. iLQR solves \eqref{eq:dt_mpc} by breaking it into a series of subproblems with first or second-order Taylor approximations of nonlinear functions $J$, $f_k$, $g_k$, and $h_k$. When combined with an augmented Lagrangian method to handle constraints (AL-iLQR), this method efficiently handles many common robotics problems \cite{Howell2019}. In this work, we deploy our quaternion MPC by modifying the open-source AL-iLQR solver ALTRO \cite{Howell2019}.

\section{Quaternion MPC}\label{sec:quat_mpc}

In this section, we present the implementation details of quaternion MPC. Specifically, we discuss the necessary modifications to the iLQR problem formulation for quaternion-based optimal control on legged robots.

\subsection{MPC Architecture}\label{sec:control}

\begin{figure}
    \vspace{2mm}
    \centering\includegraphics[width=1.0\linewidth]{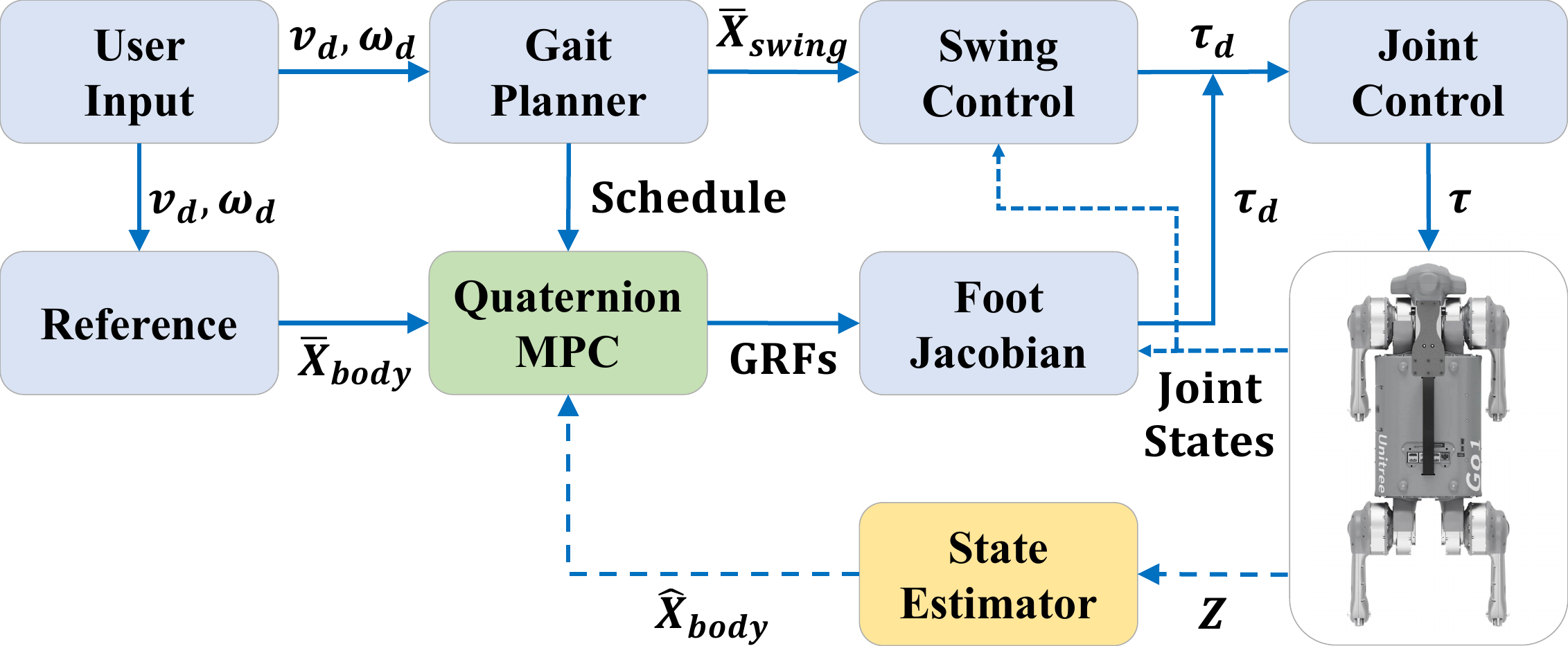}
    \caption{Quaternion MPC control architecture. The green component operates at 140 Hz, while the blue and yellow elements function at 1000 Hz.}\label{fig:control_arch}    
\end{figure}

The control architecture is illustrated in Fig. \ref{fig:control_arch}. A user inputs desired linear and angular velocities, which are treated as references in the MPC problem. A gait planner generates a foot-contact schedule, which are treated as constraints in the MPC problem. The nonlinear MPC module computes the desired ground reaction forces in each foot by solving problem \eqref{eq:dt_mpc} using a single-rigid-body model with foot contact and friction cone constraints \cite{Bledt2018}. These desired ground reaction forces are then mapped to joint torques via inverse kinematics and tracked by a low-level PD controller. For state estimation, we implemented an observability-constrained extended Kalman Filter that combines measurements from joint encoders and a body Inertial Measurement Unit (IMU) to estimate both the robot's torse pose and foot contact locations \cite{Bloesch2012}.

\subsubsection{Coordinate Systems}

Three coordinate systems are defined: \textit{World} \{\textit{W}\}, \textit{Body} \{\textit{B}\}, and \textit{Relative} \{\textit{R}\} (Fig. \ref{fig:coordinates}). The Body frame's origin is located at the CoM of the modeled single rigid body and shares the same orientation. The Relative frame shares the same origin as the Body frame but only considers the yaw angle of the body's attitude for its orientation. Commands from the user via joysticks are defined in the Relative coordinate system.


\subsubsection{SRB Dynamics}



\begin{figure}
    \centering
    \begin{minipage}[t]{0.49\linewidth}
        \centering
        \raisebox{0.2\height}{\includegraphics[width=1.0\linewidth]{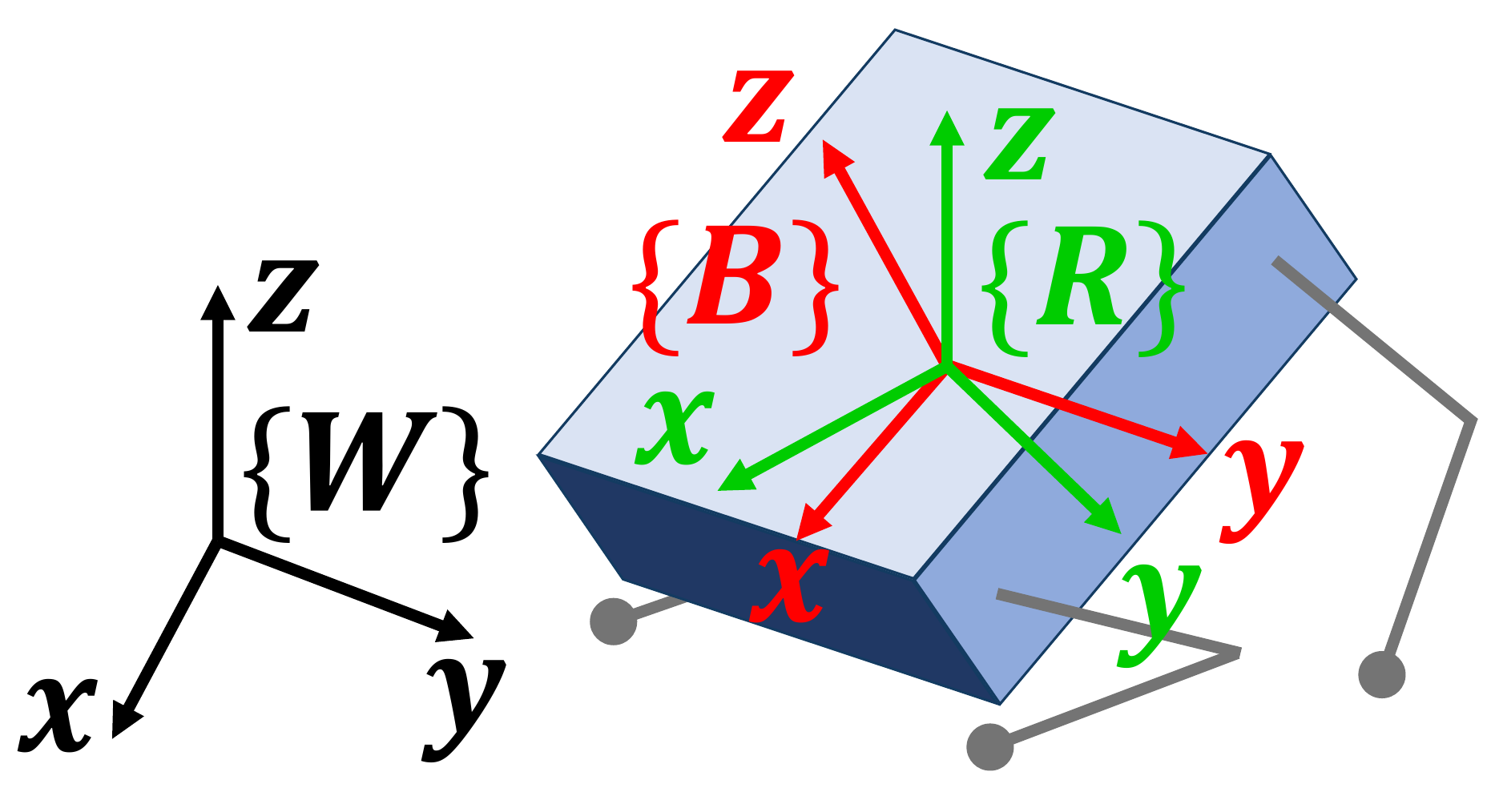}}
        \caption{Coordinate systems}\label{fig:coordinates}
    \end{minipage}%
    \begin{minipage}[t]{0.49\linewidth}
        \centering
        \includegraphics[width=1.0\linewidth]{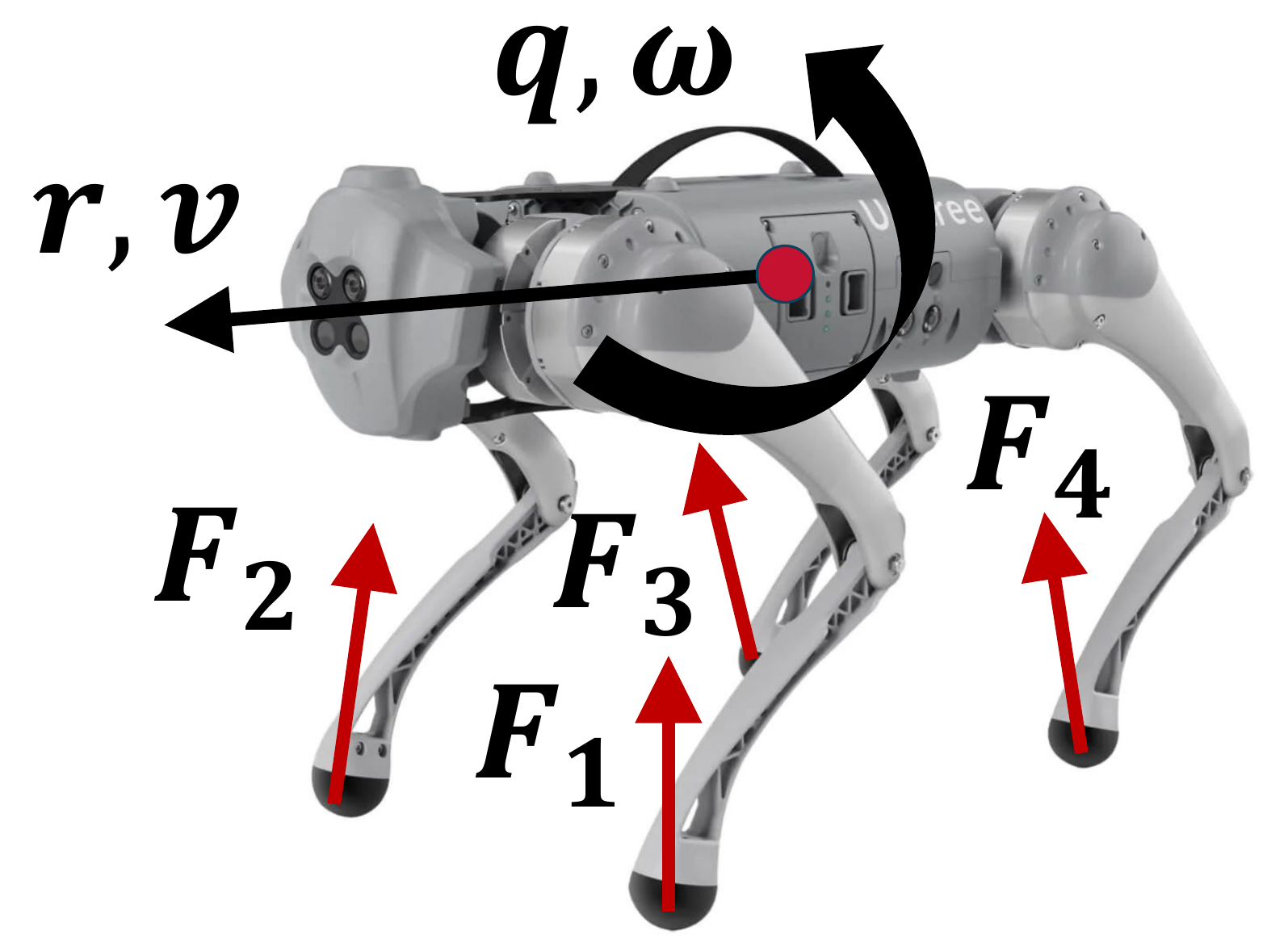}
        \caption{SRB dynamics}\label{fig:srb_dynamics}
    \end{minipage}
\end{figure}

The SRB dynamics for legged robots can be derived as follows:
\begin{equation}\label{eq:lin_acc}
    \Ddot{\textbf{r}}=\Dot{\textbf{v}}=\frac{1}{m}\sum_{i}{\textbf{F}_i},
\end{equation}
\begin{equation}\label{eq:attitude}
    \Dot{\textbf{q}}=\frac{1}{2}\textbf{q}\otimes \hat{\boldgreek{\omega}},
\end{equation}
\begin{equation}\label{eq:ang_acc}
    \textbf{M}=\textbf{I}\Dot{\boldgreek{\omega}}+\boldgreek{\omega}\times \left(\textbf{I}\boldgreek{\omega}\right),
\end{equation}
where $\textbf{r}\in \mathbb{R}^3$ denotes the position of the body, $\textbf{q}\in \mathbb{H}$ represents the body attitude in unit quaternions, $\textbf{v}\in\mathbb{R}^3$ is the linear velocity of the body, $\boldgreek{\omega}\in\mathbb{R}^3$ is the angular velocity of the body, $\textbf{F}_i\in\mathbb{R}^3$ corresponds to the external force at the $i$-th contact point, $m\in\mathbb{R}$ is the body's mass, $\textbf{I}\in\mathbb{R}^{3\times 3}$ is the body inertia matrix (Fig. \ref{fig:srb_dynamics}), and $\textbf{M}\in\mathbb{R}^3$ is the applied torques. We define the state vector $\textbf{x}\in\mathbb{R}^{13}:=\left[\textbf{r}^\intercal \text{ }\textbf{q}^\intercal \text{ }\textbf{v}^\intercal \text{ }\boldgreek{\omega}^\intercal \text{ }\right]^\intercal $ and the control vector $\textbf{u}\in\mathbb{R}^{3n_c}:=\left[\textbf{F}_1^\intercal \text{ ... }\textbf{F}_i^\intercal \right]^\intercal $, then we can get the continuous-time dynamics:
\begin{equation}\label{eq:dynamics}
    f_{cont}\left(\textbf{x}, \textbf{u}\right)=\Dot{\textbf{x}}=
    \begin{bmatrix}
        \textbf{v} \\
        \frac{1}{2}\textbf{q}\otimes \hat{\boldgreek{\omega}} \\
        \frac{1}{m}\left(\sum_{i}{\textbf{F}_i}-m\textbf{g}\right) \\
        \textbf{I}^{-1}\left[\textbf{M}-\boldgreek{\omega}\times \left(\textbf{I}\boldgreek{\omega}\right)\right]
    \end{bmatrix}
\end{equation}
All variables in Eq. \ref{eq:dynamics}, with the exception of attitude $\textbf{q}$, are represented in the Body frame.

The discrete-time dynamics function $\textbf{x}_{k+1} = f_k(\textbf{x}_k, \textbf{u}_k)$ is computed from the continuous dynamics function (Eq. \ref{eq:dynamics}) using the midpoint method.
\subsection{Modification to AL-iLQR}


In iLQR, we linearized the robot dynamics about some reference state and control trajectories, $\bar{\textbf{x}}$ and $\bar{\textbf{u}}$, with \eqref{eq:quat_fun_jac}. The linearized error state dynamics are:
\begin{equation}\label{error_dynamics}
    \delta \textbf{x}_{k+1}=\textbf{A}_k\delta \textbf{x}_k + \textbf{B}_k\delta \textbf{u}_k.
\end{equation}
where:
\begin{equation}\label{error_A}
    \textbf{A}_k=E\left(\bar{\textbf{x}}_{k+1}\right)^\intercal  \frac{\partial f}{\partial \textbf{x}} |_{\bar{\textbf{x}}_{k}, \bar{\textbf{u}}_{k}} E\left(\bar{\textbf{x}}_{k}\right),
\end{equation}
\begin{equation}\label{error_B}
    \textbf{B}_k=E\left(\bar{\textbf{x}}_{k+1}\right)^\intercal  \frac{\partial f}{\partial \textbf{u}} |_{\bar{\textbf{x}}_{k}, \bar{\textbf{u}}_{k}},
\end{equation}
where the error-state Jacobian matrix $E\left(\textbf{x}\right)\in\mathbb{R}^{13\times12}$ is:
\begin{equation}\label{error_E}
    E\left(\textbf{x}\right)=
    \begin{bmatrix}
        \textbf{I}_3 & & & \\
        & G\left(\textbf{q}\right) & & \\
        & & \textbf{I}_3 & \\
        & & & \textbf{I}_3
    \end{bmatrix}.
\end{equation}
We apply the same modifications to the second-order Taylor expansion of the "action-value function" $Q\left(x, u\right)$ in AL-iLQR \cite{Howell2019}:
\begin{equation}\label{Q_xx}
    \textbf{Q}_{xx}=\textbf{l}_{xx}+\textbf{A}_k^\intercal \textbf{P}_{k+1}\textbf{A}_k + \textbf{c}_x^\intercal \textbf{I}_\mu \textbf{c}_x,
\end{equation}
\begin{equation}\label{Q_uu}
    \textbf{Q}_{uu}=\textbf{l}_{uu}+\textbf{B}_k^\intercal \textbf{P}_{k+1}\textbf{B}_k + \textbf{c}_u^\intercal \textbf{I}_\mu \textbf{c}_u,
\end{equation}
\begin{equation}\label{Q_ux}
    \textbf{Q}_{ux}=\textbf{l}_{ux}+\textbf{B}_k^\intercal \textbf{P}_{k+1}\textbf{A}_k + \textbf{c}_u^\intercal \textbf{I}_\mu \textbf{c}_x,
\end{equation}
\begin{equation}\label{Q_x}
    \textbf{Q}_{x}=\textbf{l}_{x}+\textbf{A}_k^\intercal \textbf{p}_{k+1} + \textbf{c}_x^\intercal (\boldgreek{\uplambda}+\textbf{I}_\mu \textbf{c}),
\end{equation}
\begin{equation}\label{Q_u}
    \textbf{Q}_{u}=\textbf{l}_{u}+\textbf{B}_k^\intercal \textbf{p}_{k+1} + \textbf{c}_u^\intercal (\boldgreek{\uplambda}+\textbf{I}_\mu \textbf{c}),
\end{equation}
where $\textbf{l}_{xx}\in\mathbb{R}^{12\times12}=\partial^2 l_k/\partial \textbf{x}^2$, $\textbf{l}_{uu}\in\mathbb{R}^{3n_c\times3n_c}=\partial^2 l_k/\partial \textbf{u}^2$, $\textbf{l}_{ux}\in\mathbb{R}^{3n_c\times12}=\partial^2 l_k/\partial \textbf{x}\partial \textbf{u}$, $\textbf{l}_{x}\in\mathbb{R}^{12}=\partial l_k/\partial \textbf{x}$, $\textbf{l}_{u}\in\mathbb{R}^{3n_c}=\partial l_k/\partial \textbf{u}$, $\textbf{P}\in\mathbb{R}^{12\times12}$ and $\textbf{p}\in\mathbb{R}^{12}$. The last term in each equation is related to the penalty term and the Lagrange multipliers for the constraint functions.

\subsection{Objective Function}

The quaternion terms play a crucial role in the objective function. To design an effective objective for a reference-tracking problem, it's common to employ quadratic costs, aiming to minimize the weighted 2-norm distance to the reference trajectory. However, the distance between unit quaternions is not accurately captured by the 2-norm because they lie on the surface of a 4D hypersphere. Thus we use a distance that is easy to compute and monotonic with the geodesic distance:
\begin{equation}\label{quat_obj}
    l_{k, q}=1-|\bar{\textbf{q}}_k^\intercal \textbf{q}_k|.
\end{equation}
Using the quaternion calculus techniques introduced in Sec. \ref{sec:background}, we can obtain the gradient and Hessian of $l_{k, q}$:
\begin{equation}\label{quat_obj_gradient}
    \nabla l_{k, q}\in\mathbb{R}^3=-\text{sign}\left(\bar{\textbf{q}}_k^\intercal \textbf{q}_k\right)\bar{\textbf{q}}_k^\intercal G\left(\textbf{q}_k\right),
\end{equation}
\begin{equation}\label{quat_obj_hessian}
    \nabla^2 l_{k, q}\in\mathbb{R}^{3\times3}=\text{sign}\left(\bar{\textbf{q}}_k^\intercal \textbf{q}_k\right)\textbf{I}_3\left(\bar{\textbf{q}}_k^\intercal \textbf{q}_k\right).
\end{equation}


\section{Experiments and Results}\label{sec:results}
This section presents the results of several legged-robot control experiments evaluating quaternion MPC:
We demonstrated quadruped dynamic walking and vertical standing between two walls, both in simulation and on hardware. In addition, we compared quaternion MPC and Euler MPC for airborne attitude control of a falling quadruped equipped with reaction wheels. We also achieved large attitude adjustments of a humanoid robot while standing.


\subsection{Experiments Setup}
We validated quaternion MPC on a Unitree Go1 robot on hardware and in simulation, and the MIT humanoid robot \cite{chignoli2021mit} in simulation. The whole control architecture, shown in Fig. \ref{fig:control_arch}, was implemented in C++ using ROS. We deployed an asynchronous multi-threaded approach for each module: the nonlinear MPC operated at $140$ Hz, while all other components, including the state estimator, ran at $1000$ Hz. For all experiments, we warm-started the nonlinear MPC with an initial control by equally distributing the body weight to all feet. For the remaining MPC solves, we warm-started with the previous solution.

All experiments were conducted on a desktop equipped with an Intel i9-12900KS CPU and 64 GB RAM. We used the Open Dynamics Engine for all simulation experiments and a motion capture system to measure robot poses on hardware.  



\subsection{Dynamic Quadruped Walking with Attitude Control}
\begin{figure}
    \vspace{2mm}
    \centering
    \resizebox{1.0\linewidth}{!}{\input{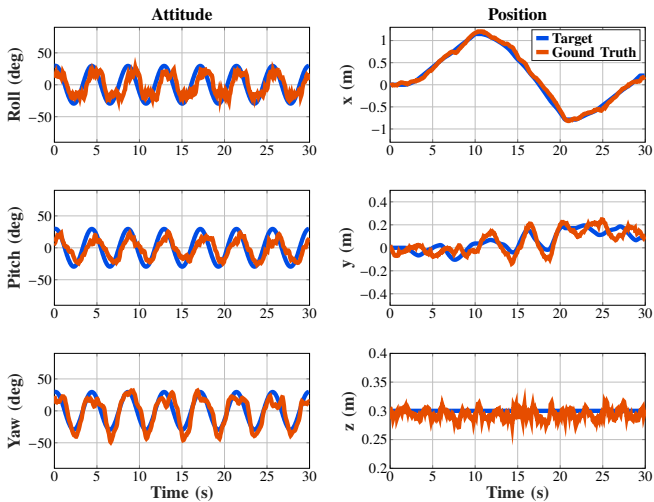}}
    \caption{Unitree Go1 position and attitude data on hardware. Quaternion MPC tracks a sinusoidal desired attitude trajectory during dynamic locomotion.}
    \label{fig:atti_walk_test}
\end{figure}

Our controller successfully enables real-world dynamic walking while maintaining full attitude control.
In this experiment, we controlled the robot to walk around in a trotting gait, while continuously sending sinusoidal angular velocity commands, with an amplitude of $30$ deg/s and a period of $4.25$ s in all three directions. The performance of the tracking control target on hardware is plotted in Fig. \ref{fig:atti_walk_test}. Our method can reliably control the robot's attitude and position with small tracking error.

Note that while input to our controller is the desired linear or angular velocity, the reference position and attitude are computed by forward integration of the current state and desired velocities. We use a planning horizon of $0.36$ s with a time discretization of $0.01$ s. 


\begin{figure}
    \centering
    \resizebox{1.0\linewidth}{!}{\input{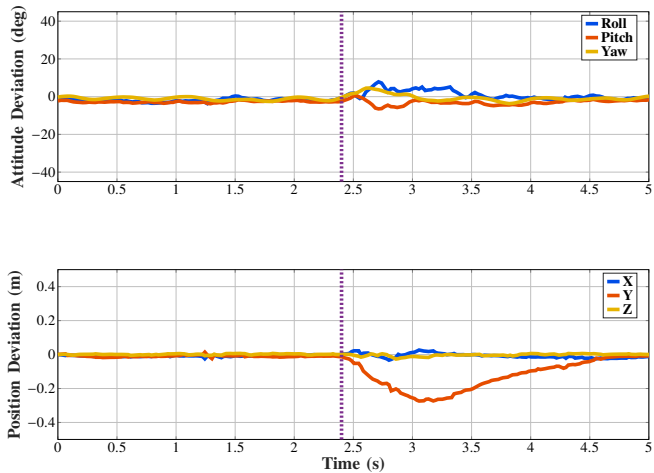}}
    \caption{Unitre Go1 position and attitude data from a hardware disturbance rejection experiment during trotting. The purple line indicates the instance when the external disturbance is applied.}
    \label{fig:kick_test_data}
\end{figure}

Additionally, we validated the robustness of our controller by disturbing the robot's torso during trotting. The deviations in attitude and position throughout this process are displayed in Fig. \ref{fig:kick_test_data}, where the purple line indicates the moment the disturbance occurs. The attitude exhibited small errors after the disturbance was applied and the position deviation recovered to approximately zero within about $2$ s.


\subsection{Standing Between Two Walls}
\begin{figure}
    \vspace{2mm}
    \centering
    \resizebox{1.0\linewidth}{!}{\input{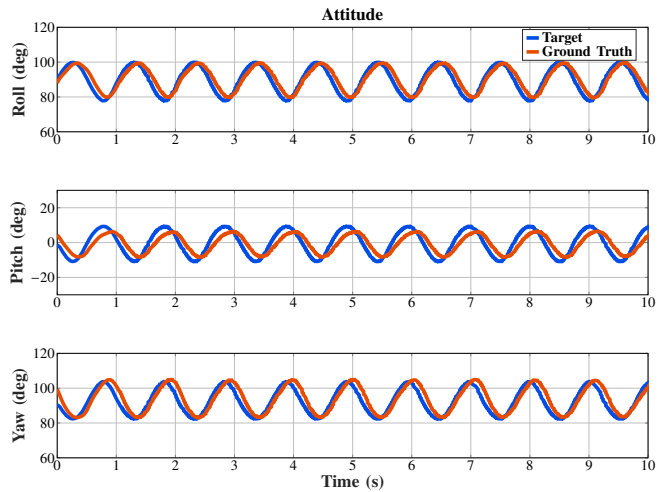}}
    \caption{Unitree Go1 attitude data from Gazebo simulation. Quaternion MPC tracks the sinusoidal desired attitude while standing vertically between two walls, Fig. \ref{fig:wall_standing}.}
    \label{fig:wall_standing_gazebo}
\end{figure}

Inspired by human rock climbers, we designed an experiment in which a quadruped robot stands vertically between two walls in simulation, an attitude configuration that results in a $90$-degree pitch angle with conventional Euler angles. Similar to the previous experiment, we applied sinusoidal angular velocity commands and our controller successfully tracked the target attitude trajectory (Fig. \ref{fig:wall_standing_gazebo}). Note that these attitudes lead to the ZYX Euler-angle-based controller failing due to singularity near the $90$-degree pitch angle. 

We also performed this experiment on hardware (Fig. \ref{fig:wall_standing}). Our method successfully enabled the Unitree Go1 robot to stand between two walls completely independent of additional weight support. However, motor thermal limits prevented us from performing meaningful attitude adjustment experiments. 




\begin{figure}
    \centering
    \begin{overpic}[trim=8cm 21cm 8cm 0cm, clip, height=4.2cm]{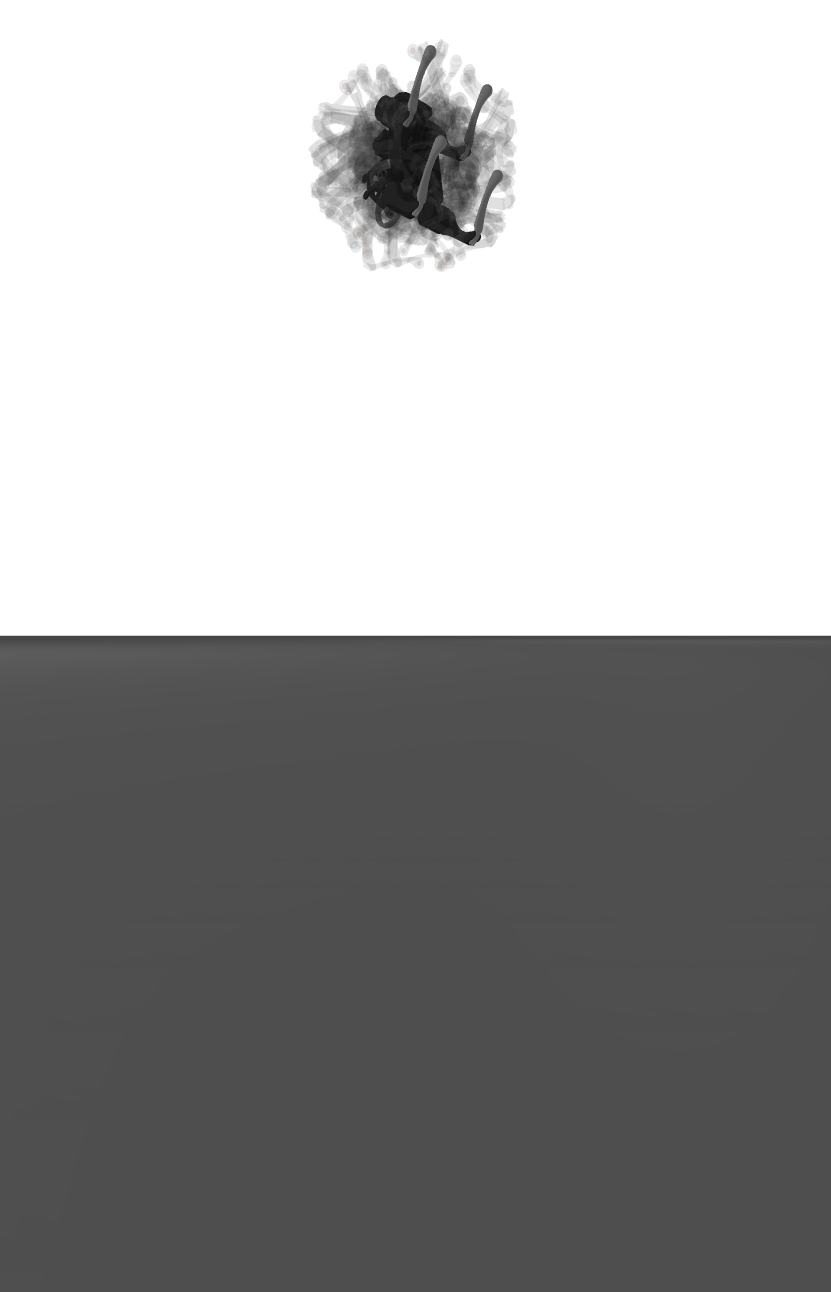}
        \put(13,115){$t=0.0$s}
    \end{overpic}
    \hfill
    \begin{overpic}[trim=8cm 21cm 8cm 0cm, clip, height=3.75cm]{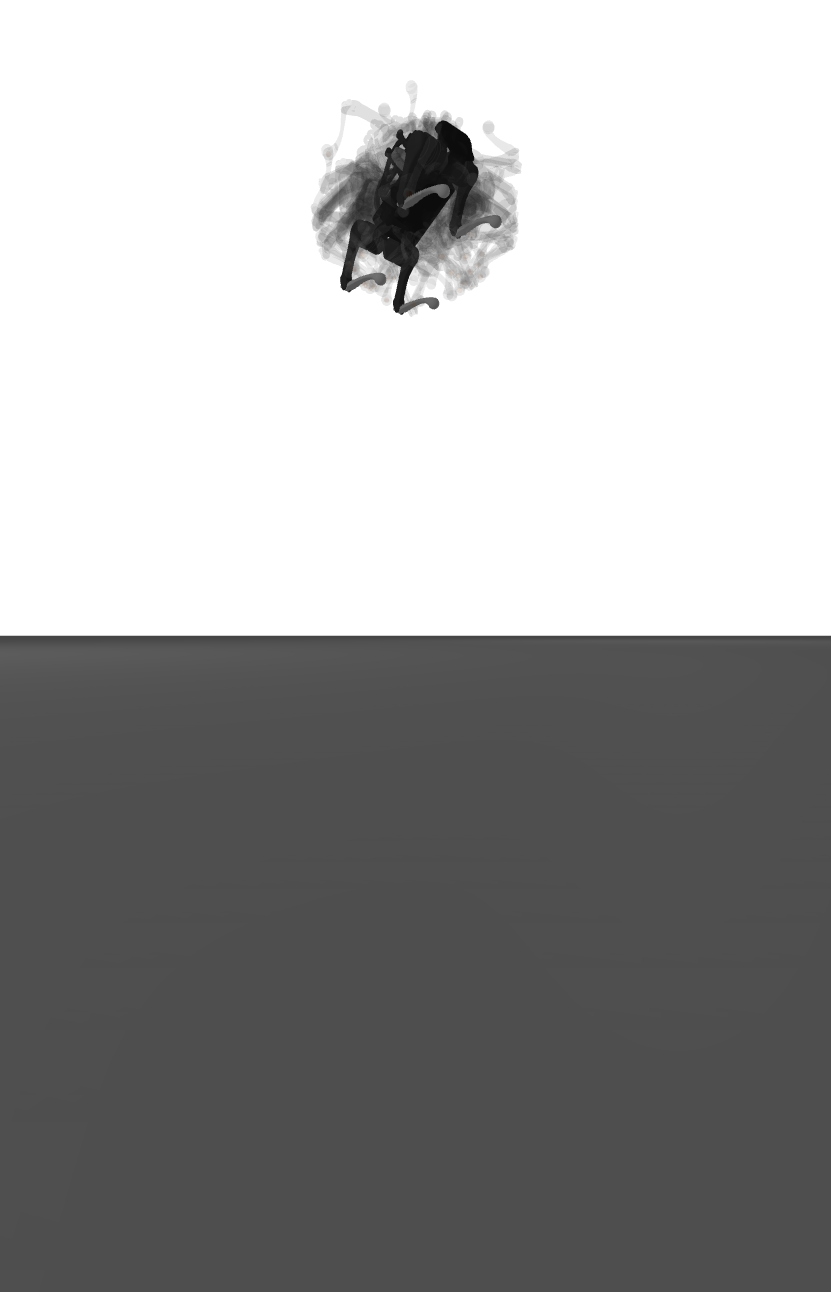}
        \put(13,115){$t=0.2$s}
    \end{overpic}
    \hfill
    \begin{overpic}[trim=8cm 21cm 8cm 0cm, clip, height=3.75cm]{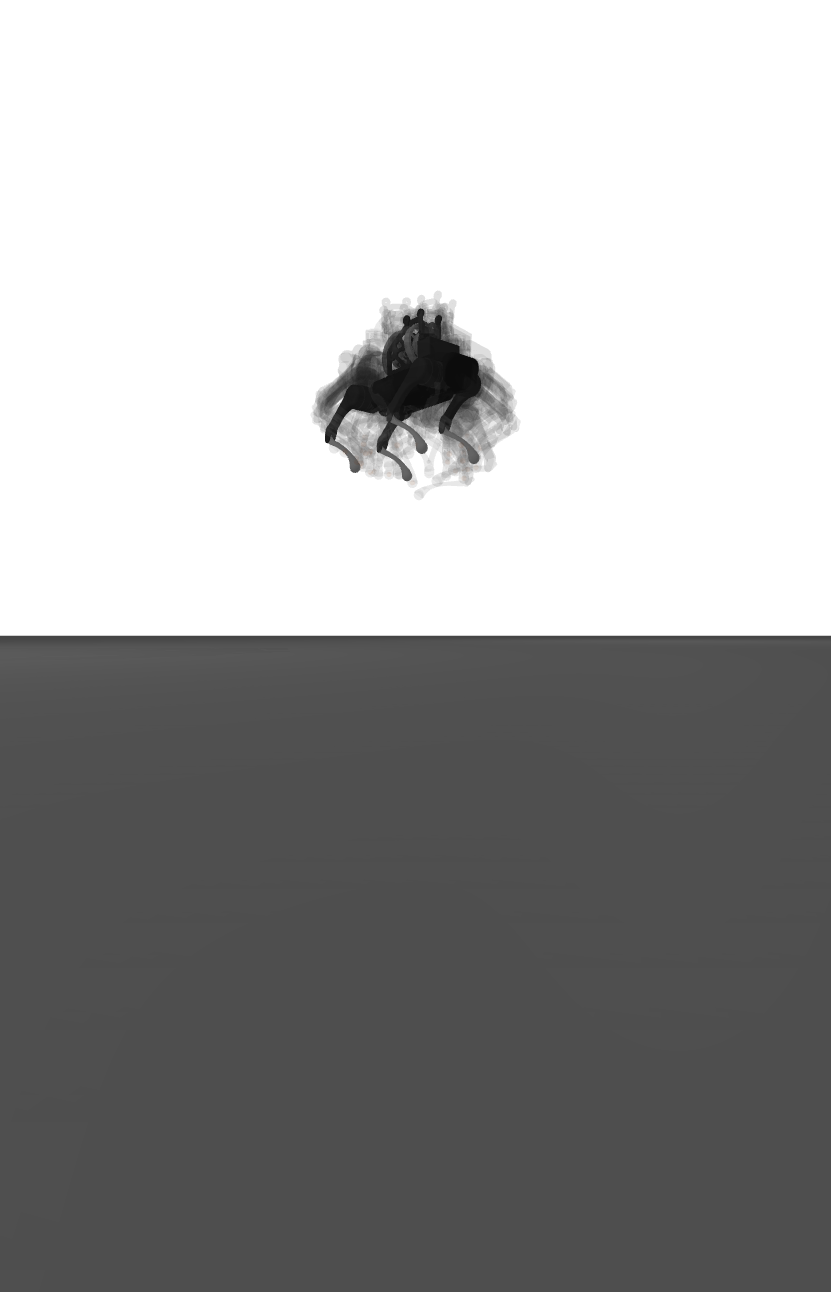}
        \put(13,115){$t=0.5$s}
    \end{overpic}
    \hfill
    \begin{overpic}[trim=8cm 21cm 8cm 0cm, clip, height=3.75cm]{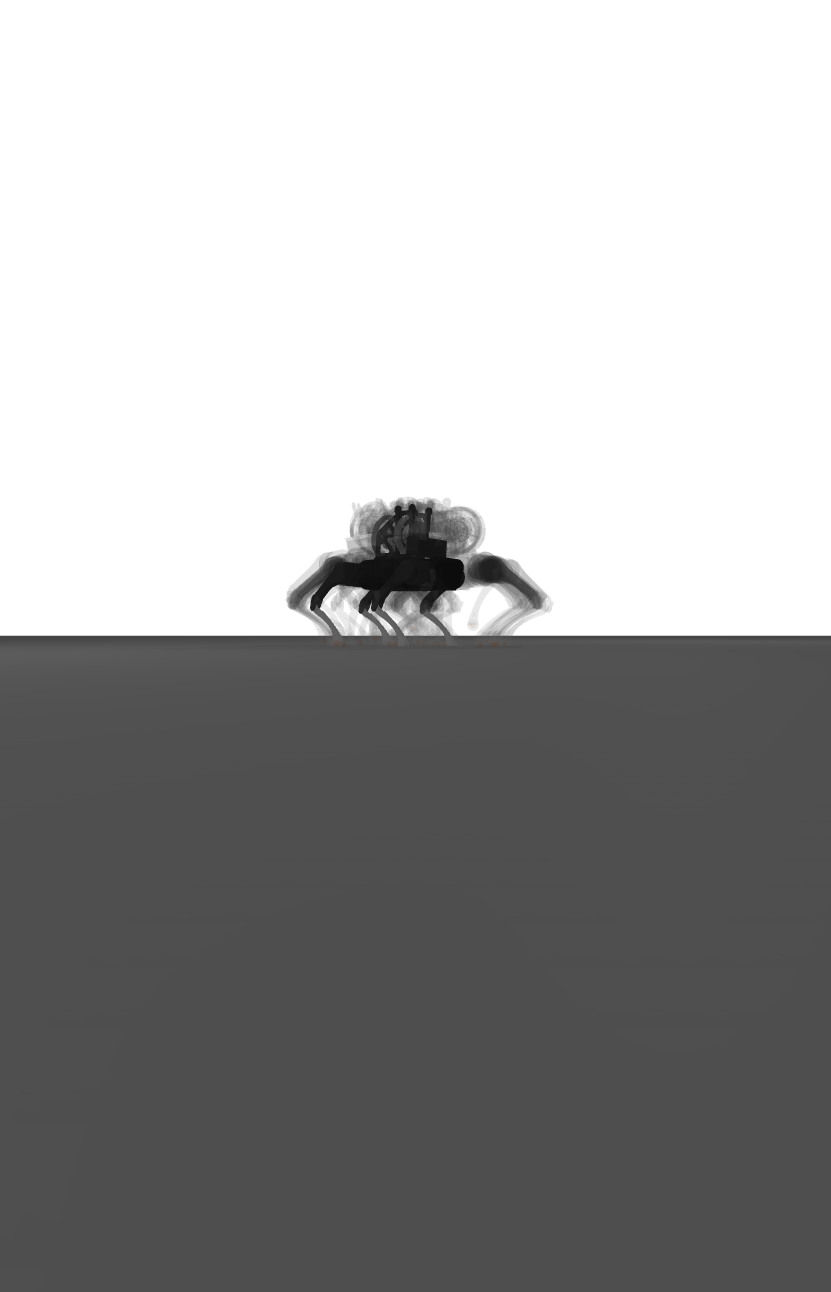}
        \put(13,115){$t=1.3$s}
    \end{overpic}

    \caption{Monte Carlo simulation of quadruped robots with roll and pitch reaction wheels. Robots were initialized from random initial attitudes from 1.75 m above ground and successfully landed on their feet 95\% percent of the time. Euler MPC only achieved 41\% success rate.}
    \label{fig:falling_cat_monte_carlo}
\end{figure}

\subsection{Airborne Attitude Control}

The ability to adjust attitude during airborne phases can be important for legged robots operating in the field. Cats are famously able to land on their feet during a fall regardless of their initial orientation \cite{Montgomery1993GaugeTO}. While humans and humanoid robots can perform similar reorientation maneuvers using their limbs, standard quadruped robots today cannot efficiently achieve this due to their light legs and lack of a flexible spine. To address this issue, researchers have adding reaction wheels \cite{Lee2023} or tails \cite{johnson_tail} to quadruped robots.

We utilized quaternion MPC to achieve stable airborne attitude control on a quadruped robot equipped with roll and pitch reaction wheels \cite{Lee2023}. Note that when using any 3-parameter rotation representation, solving this problem is inherently prone to singularities. To evaluate our controller, we conducted Monte Carlo simulations of ``falling cat'' experiments with our reaction-wheel-equiped quadruped robot (Fig. \ref{fig:falling_cat_monte_carlo}). In each trial, the initial state of the robot was 1.75 meters above the ground with a randomized initial attitude $\textbf{q}_0$. The target attitude $\bar{\textbf{q}}$ was obtained by solving the following optimization problem:
\begin{equation}\label{eq}
    \begin{array}{ll}
        \underset{\bar{\textbf{q}}}{\mbox{minimize}} & J = 1 - |\bar{\textbf{q}}^\intercal \textbf{q}_0|\\
        \mbox{subject to}
        & \bar{q}_x = \bar{q}_y = 0,\\
        & \bar{q}_z^2 + q_w^2 = 1.
    \end{array}
\end{equation}
By minimizing the aforementioned metric (Eq. \ref{quat_obj}) between the initial and target attitudes, while ensuring zero roll and pitch at landing, we can determine a feasible landing attitude that is closest to the initial attitude.

The MPC directly outputs torque commands to the reaction wheel motors, eliminating the need for a low-level controller. 
Over $100$ trials, we achieved a success rate of approximately $95\%$.

We also performed comparison experiments using Euler MPC. The full nonlinear Euler's equations were used in the dynamics without any approximation. The same method was used to generate the initial and the target attitudes. Euler MPC achieved a much lower 41\% success rate over 100 trials.

\subsection{Humanoid Attitude Control}

\begin{figure}
    \vspace{2mm}
    \centering\includegraphics[width=1.0\linewidth]{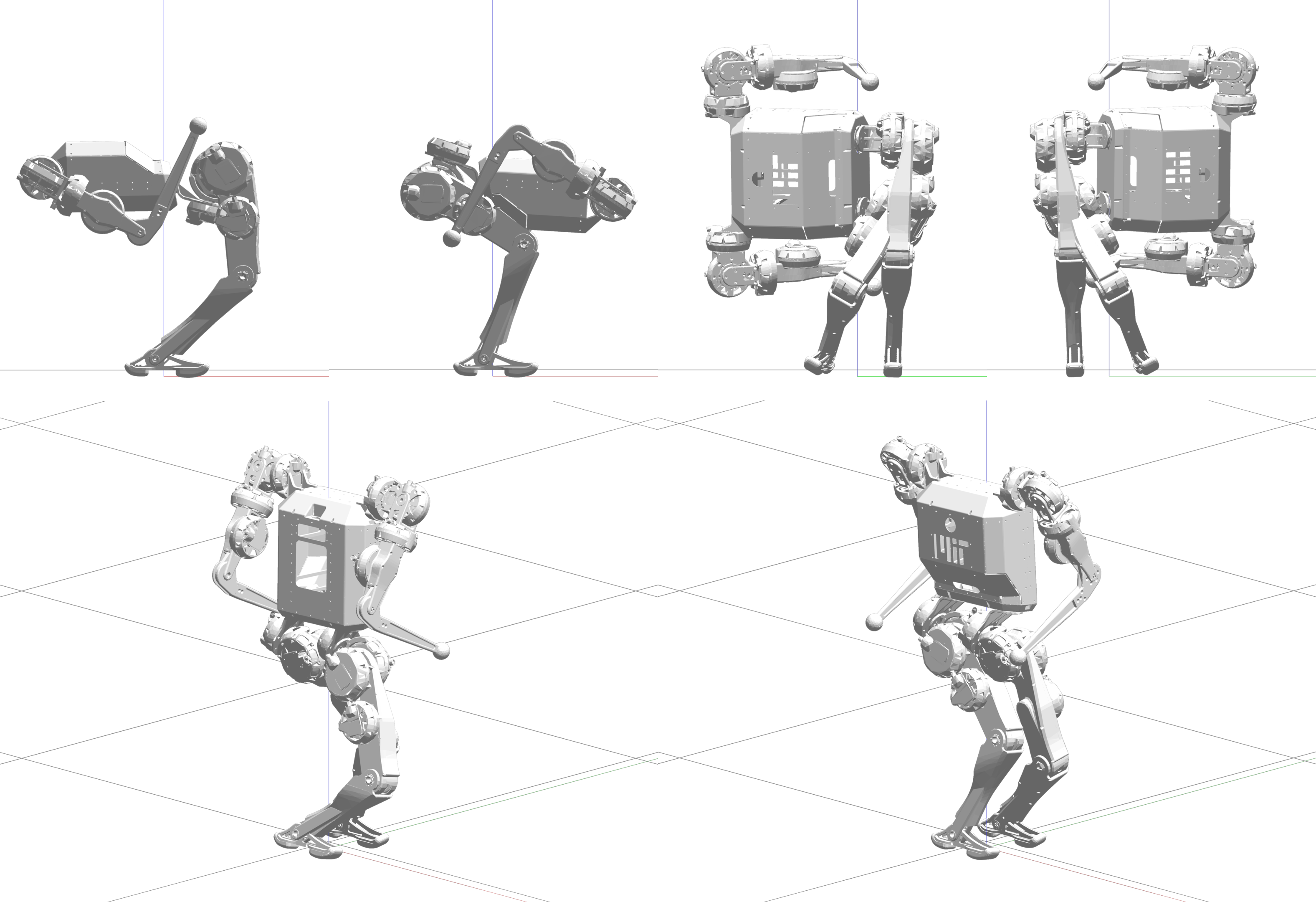}
    \caption{Singularity-free attitude control of the MIT humanoid robot \cite{chignoli2021mit} with quaternion MPC.}\label{fig:humanoid}
\end{figure}

We also applied quaternion MPC to a humanoid robot, where singularity-free attitude control is much more critical during locomotion and manipulation tasks. We evaluated our method in simulation on the MIT Humanoid standing on two feet (Fig. \ref{fig:humanoid}) where we controlled the torso attitude in three directions independently, each with a range of more than 180 degrees. Quaternion MPC effectively balanced the robot during large attitude changes (Fig. \ref{fig:humanoid}). Note that, although we did not test the locomotion capabilities of quaternion MPC in this experiment, we used the same floating-base single-rigid-body MPC formulation and control architecture described in Sec. \ref{sec:quat_mpc}. We used a planning horizon of $0.36$ s with a time discretization of $10$ ms.

\begin{figure}
    \vspace{2mm}
    \centering
    \resizebox{1.0\linewidth}{!}{\input{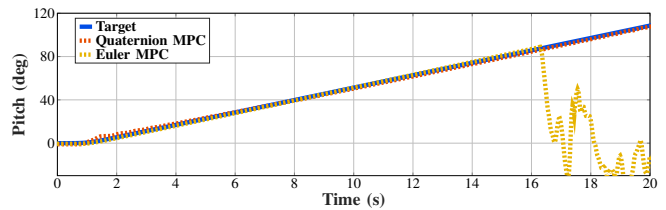}}
    \caption{Comparison between quaternion MPC (red) and Euler MPC (yellow) during a pitch angle adjustment on a simulated MIT humanoid robot. Quaternion MPC successfully tracks the desired attitude (blue) over the entire range of motion, while Euler MPC fails as the pitch angle approaches a singularity near $90$ degrees. 
    }
    \label{fig:humanoid_mpc_compare}
\end{figure}

We compared quaternion MPC to Euler MPC \cite{chignoli2021mit}, Fig. \ref{fig:humanoid_mpc_compare}. As expected, Euler MPC failed when the pitch angle approached 90 degrees while quaternion MPC successfully tracked the entire desired attitude range (Fig. \ref{fig:humanoid_mpc_compare}).

\section{Conclusions}\label{sec:conclusions}

We have presented quaternion MPC, a singularity-free nonlinear MPC framework for legged robots using an SRB model with quaternions. We demonstrate the capabilities of our controller on a range of hardware and simulation experiments that undergo significant attitude changes for both quadruped and humanoid robot platforms. Future works include incorporating the quaternion attitude representation for whole-body MPC formulations and hardware implementations of our controller on real-world humanoid robot platforms. 




\bibliographystyle{IEEEtran}
\bibliography{references}

\end{document}